\def\CFSPreprintMode{1}
\def\CFSEmbedSupplement{1}
\documentclass{article}
\PassOptionsToPackage{numbers}{natbib}
\usepackage[preprint]{neurips_2026}
\makeatletter
\renewcommand{\@notice}{}
\makeatother
\usepackage{placeins}
\usepackage{multirow}
\usepackage{graphicx}
\usepackage{float}
\usepackage{subcaption}
\usepackage{amsmath,amssymb,amsthm}

\newtheorem{proposition}{Proposition}

\newcommand{\zvec}{\mathbf{z}}

\newcommand{\dhat}{\hat{\mathbf{d}}}

\newcommand{\mur}{\boldsymbol{\mu}_r}
\newcommand{\R}{\mathbb{R}}
\usepackage[utf8]{inputenc}
\usepackage[T1]{fontenc}
\usepackage{hyperref}
\usepackage{url}
\usepackage{booktabs}
\usepackage{amsfonts}
\usepackage{nicefrac}
\usepackage{microtype}
\usepackage{xcolor}
\usepackage{soul}
\usepackage{xr}
\ifdefined\CFSEmbedSupplement
\else
\fi

\title{Counterfactual Feature Steering for Adjacency-Preserving Identity Unlearning}
\date{}
\ifdefined\CFSPreprintMode
  \author{
  FNU Shivam \\
  West Virginia University \\
  Morgantown, WV, USA
  \And
  Layth~Sliman \\
  EFREI Paris \\
  Villejuif, France
  \And
  Ramana~Reddy \\
  West Virginia University \\
  Morgantown, WV, USA
  \And
  Donald~Adjeroh \\
  West Virginia University \\
  Morgantown, WV, USA
  \And
  Binod~Bhattarai \\
  University of Aberdeen \\
  Aberdeen, United Kingdom
  \And
  Prashnna~Gyawali \\
  West Virginia University \\
  Morgantown, WV, USA
}
\else
  \author{Anonymous Authors}
\fi

\begin{document}

\maketitle

\begin{abstract}
Identity unlearning in text-to-image diffusion models is increasingly important for privacy, safety, and ethical deployment, as it aims to remove the model's ability to generate specific individuals.
In this work, we study identity unlearning with \emph{adjacency preservation}, where the goal is to remove a target individual's identity while preserving related non-target identities. We propose Counterfactual Feature Steering (CFS), a method requiring one-time SAE training that performs bounded inference-time concept editing in the latent space of a sparse autoencoder attached to the text encoder. CFS derives a target direction by contrasting forget and retain distributions, then applies a per-token projection proportional to each token's current alignment with this direction, yielding counterfactual representations that suppress the target while limiting collateral damage. To evaluate adjacency preservation, we introduce the Adjacency-Weighted Unlearning Score (AWUS), which measures the tradeoff between target erasure and collateral damage, and extend it to face-specific identity similarity using ArcFace. 
Across 4,000 prompts derived from Celeb-100, spanning 100 celebrity identities, CFS achieves the strongest evaluated tradeoff between identity erasure
of target identities and preservation of adjacent and generic concepts, while remaining stable across intervention strengths where other steering-based methods degrade. Independent ArcFace evaluation and a 10,000-caption COCO study further examine identity and general preservation. Quantitative nudity-suppression benchmarks and qualitative artistic-style examples provide evidence of transfer beyond identities on SD~1.4.


\end{abstract}

\section{Introduction}
\label{sec:intro}

Generative models have recently shown striking abilities across domains \citep{rombach2022high, radford2021learning}, from natural language understanding to high-fidelity image synthesis. Text-to-image models \citep{rombach2022high}, in particular, can translate simple prompts into rich visual content, but this capability also raises safety and privacy concerns: models trained on large web-scale datasets may reproduce sensitive, copyrighted, or personal data without consent. Mitigating these risks is important for both safety and legal compliance, including regulations such as 
European Union's (EU) General Data Protection Regulation (GDPR) \citep{regulation2016regulation}. As a result, targeted model editing and unlearning have received increasing attention as ways to remove undesirable concepts while preserving broader model utility \citep{ginart2019making, gandikota2023erasing}.

Toward removing such undesirable concepts, a key challenge is first to locate them within the model’s internal representations. Mechanistic interpretability approaches—particularly Sparse Autoencoders (SAEs) \citep{bricken2023towards, huben2024sparse} offer a promising avenue for this \citep{cywinski2025saeuron, pach2025sparse, kamenetsky2025saedit}. SAEs are trained to decompose dense model activations into a set of sparse features, often aligning with more interpretable, monosemantic concepts \citep{surkov2024one, cywinski2025saeuron}. By operating in this sparse feature space, one can identify components associated with specific attributes (e.g., nudity or identity-related cues) and apply targeted interventions, such as penalizing or steering their activations. This enables selective unlearning or editing of concepts while minimizing unintended effects on the model’s broader capabilities \citep{kim2025concept, cywinski2025saeuron, wang2025model}.

While such steering-based approaches are appealing because they are lightweight and do not require access to training data or model retraining, existing methods typically apply a fixed intervention strength in the latent space to steer the model away from a target concept. This leads to two recurring structural issues. First, this strength is \emph{unbounded}: increasing it often has limited impact up to a point before abruptly degrading outputs or causing representation collapse. Second, because the same strength is applied uniformly across prompts and tokens, the intervention may fail to distinguish cases where the target concept is strongly present from those where it is only weakly expressed, increasing the risk of unintended changes to semantically adjacent concepts.

This gives rise to the central problem we study in this paper: fine-grained unlearning, where the goal is not only to remove a target concept, but to do so while preserving semantically adjacent concepts. This distinction is especially important for identity unlearning. For example, removing one high-profile technology CEO should not cause the model to also suppress other visually or semantically related public figures because they occupy nearby regions in representation space. When identities share overlapping attributes, such as profession, social context, appearance, or prompt distribution—unlearning methods must avoid treating these shared features as part of the target concept. We refer to this challenge as \emph{adjacency preservation}: the ability to erase a specified concept while maintaining the model’s ability to generate related but non-target concepts.
Prior work has already investigated related collateral-erasure problems: FADE formalizes concept neighborhoods, while EraseBench measures spillover degradation across non-target concepts \citep{thakral2025fine, amara2025erasebench}. We focus here on input-adaptive, inference-time identity suppression and evaluate whether it can preserve neighboring identities.

To address this gap, we take a different approach to concept steering. Rather than pushing representations away from a target concept in an unconstrained manner, we first measure how strongly the current activation aligns with that concept and then apply a controlled correction.
Based on this idea, we propose \textbf{Counterfactual Feature Steering (CFS)}, a method requiring one-time domain-level SAE training that performs bounded, input-adaptive projections in the latent space of a SAE attached to the text encoder. 
For each token, CFS estimates a scalar signal reflecting alignment with the target concept and applies a projection step that reduces this signal by moving the representation toward a retain distribution, while keeping the change minimal. 
This formulation yields several properties along the selected target axis: the intervention cannot cross the retain-side boundary for $s\in[0,1]$, and it is selective and adaptive because tokens with nonpositive target alignment remain unchanged. At $s=1$ it is the minimum-Euclidean-norm projection onto the retain half-space in SAE coordinates; these geometric properties do not themselves guarantee preservation of all neighboring concepts or image attributes. 

We evaluate CFS primarily in the context of identity unlearning, with a particular focus on preserving semantically adjacent concepts. To this end, we conduct experiments on a text-to-image generative model, Stable Diffusion, where we apply CFS to remove identities from the Celeb-100 dataset and analyze its effect on related identities (e.g., unlearning Barack Obama and evaluating its impact on Joe Biden). Since identity-level unlearning with explicit \emph{adjacency preservation} has not been systematically studied, we introduce a suite of metrics to quantify overall unlearning utility. In particular, we measure target erasure while minimizing collateral damage to adjacent concepts. These effects are evaluated across multiple spaces, including representation space using CLIP, face embedding space using ArcFace, and perceptual similarity using LPIPS. Beyond identity unlearning, we evaluate CFS on nudity removal and provide qualitative style-editing examples. Our intended deployment is a hosted inference setting where a model provider controls the intervention and can activate target-specific directions without maintaining a separate weight-edited checkpoint; it is not a tamper-resistant removal mechanism for users controlling the serving code.
We summarize our contributions as follows:
\begin{itemize}
    \item We identify \emph{adjacency preservation} as a key challenge in fine-grained unlearning, showing that removing a target concept can inadvertently affect semantically related non-target concepts.
\item We propose \textbf{Counterfactual Feature Steering (CFS)}, an SAE-based steering method with one-time domain-level training and bounded, input-adaptive projections in latent space.
\item We introduce adjacency-aware evaluation metrics for identity unlearning that jointly measure target erasure and collateral damage across CLIP, ArcFace, and LPIPS spaces.
\item We demonstrate that CFS improves fine-grained identity unlearning while better preserving adjacent identities, and provide quantitative nudity-suppression and qualitative style-editing.
\end{itemize}

\section{Related Work}
\label{sec:related}

\paragraph{Concept unlearning in diffusion models.}
The problem of removing a concept from a pretrained diffusion model
without full retraining has been approached from two directions.
\emph{Weight-editing methods} modify model parameters to suppress a
target concept. ESD~\citep{gandikota2023erasing} fine-tunes the UNet
cross-attention weights using a contrastive objective that pushes
outputs away from the concept. UCE~\citep{gandikota2024unified} derives
a closed-form weight update that preserves manually specified anchor
concepts. MACE~\citep{lu2024mace} combines LoRA fine-tuning with
segmentation-guided masks for mass concept erasure across many targets
simultaneously. These methods are effective but require modifying and
storing updated weights, making them slow and difficult to reverse or compose.

In contrast, \emph{inference-time steering methods} operate on intermediate
activations without modifying weights, making them lightweight and
reversible. SLD~\citep{schramowski2023safe} suppresses unsafe content
by modifying the classifier-free guidance signal at each diffusion
step. One Step Is Enough~\citep{surkov2024one} applies a fixed
mean concept vector to the UNet residual stream of SDXL-Turbo.
SAeUron~\citep{cywinski2025saeuron} identifies and multiplicatively
suppresses concept-related SAE features in the UNet activations.
Concept Steerers~\citep{kim2025concept} uses a sparse
autoencoder trained on the CLIP text encoder to identify concept
directions and subtracts them at inference time.
FADE~\citep{thakral2025fine} formalizes a Concept Neighborhood to preserve related concepts during erasure, and EraseBench~\citep{amara2025erasebench} documents collateral degradation of visually and semantically related concepts. CFS differs in its inference-time, input-dependent projection mechanism and identity-specific preservation analysis.

\paragraph{Sparse autoencoders for mechanistic interpretability.}
Sparse autoencoders have emerged as a tool for decomposing dense neural
network activations into interpretable, approximately monosemantic
features~\citep{bricken2023towards,
templeton2024scaling}. Originally developed for language models \citep{huben2024sparse}, they
have been applied to diffusion model for concept
identification~\citep{tian2025sparse, tinaz2025emergence, kim2025concept, cywinski2025saeuron, surkov2024one}.

\paragraph{Machine unlearning beyond diffusion models.}
Machine unlearning~\citep{cao2015towards, ginart2019making} originally targeted
classification models with the goal of removing the influence of
specific training samples. More recent work addresses unlearning in
large language models~\citep{eldan2023s, yao2024large} and
biometric recognition systems~\citep{shivam2025cure}. In contrast, generative unlearning requires removing a distributed semantic concept that is not tied to a single data point but encoded across many prompts and model parameters, making the problem
fundamentally different.

\section{Counterfactual Feature Steering}
\label{sec:method}


Given a prompt, we construct a counterfactual representation by projecting dense activations from an internal layer of the text encoder into an SAE latent space. We then intervene on selected sparse features to reduce alignment with target concepts and decode the modified representation back into the model. 
This yields a bounded, input-dependent update: the intervention is applied
only when the current SAE activation aligns with the target-concept direction,
thereby suppressing the target concept while preserving unrelated information.

\begin{figure*}[t]
\centering
\includegraphics[width=\textwidth]{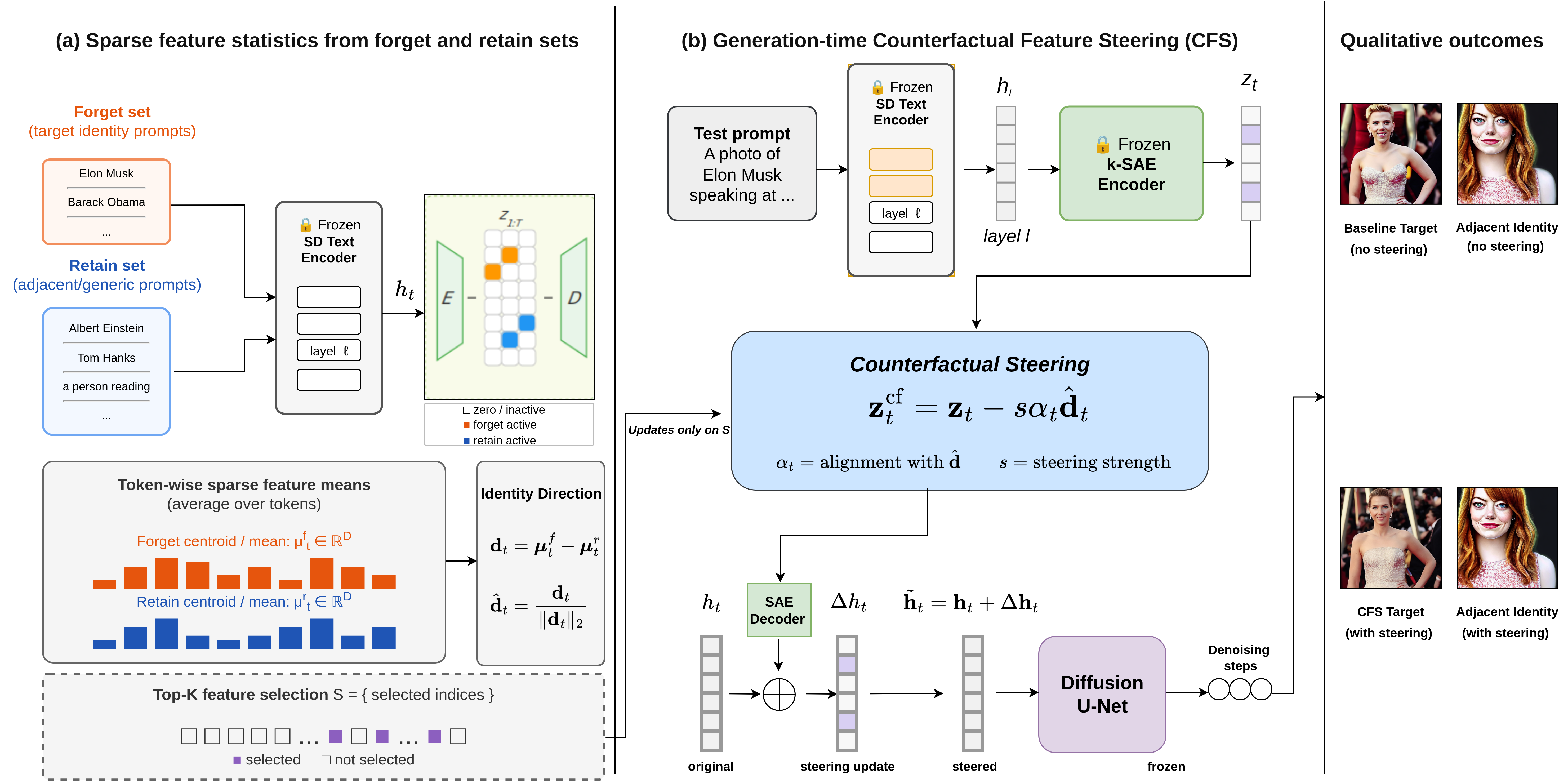}
\caption{%
  \textbf{Counterfactual Feature Steering (CFS).}
  \textbf{(a)} CFS computes sparse forget/retain statistics with a frozen
  K-SAE and derives an identity direction.
  \textbf{(b)} During generation, CFS projects the sparse code toward the
  retain distribution, decodes the sparse edit back to hidden space, and
  injects the resulting update into the frozen diffusion U-Net. The target
  identity is removed while adjacent and generic concepts are preserved.
}
\vspace{-1.2em}
\label{fig:method}
\end{figure*}

Our method consists of two main components: (i) identifying a direction in feature space that captures the target concept while avoiding features associated with retained concepts, and (ii) applying a per-token projection at inference time that edits SAE features along this direction to produce counterfactual representations. A schematic overview of the method is shown in Fig.~\ref{fig:method}.


\subsection{Preliminaries}
\label{sec:prelim}

\paragraph{Text-conditional diffusion.}
Stable Diffusion~1.4 generates an image from a prompt $p$ by first tokenising 
the prompt into a fixed-length sequence of $T = 77$ tokens (padding or truncating 
as necessary) and then encoding this sequence with a CLIP text encoder to 
produce hidden states $\mathbf{h}_{1:T} \in \mathbb{R}^{T \times d}$. The value 
$T = 77$ is determined by CLIP's maximum context length~\cite{radford2021learning}.
These embeddings condition a U-Net through cross-attention, and classifier-free 
guidance combines conditional and unconditional branches. We intervene at layer~$\ell\!=\!9$ of the CLIP text encoder.
We select this layer by ablation (Section~\ref{sec:abl}):
layer~9 achieves the highest observed AWUS among layers tested (4, 6, 8, 9, 10, 11). This is a task- and backbone-specific empirical choice, not a universal layer-selection.

\paragraph{Sparse autoencoder representation.}
We train a k-sparse autoencoder (k-SAE)~\citep{makhzani2013k} on layer-$\ell$
text-encoder activations~\citep{bricken2023towards,huben2024sparse} collected from the full prompt corpus. The identity-domain SAE is trained once (approximately 2 hours in our reported setup) and reused across all eight identity targets; new targets require only cacheable no-gradient forget/retain statistics and do not require training a separate SAE or editing diffusion-model weights. The nudity and style experiments use their own reusable domain-level SAEs. Per-target cached statistics occupy approximately 925~KB. Computational accounting is detailed in Supplementary Appendix~\ref{app:runtime}.
Each hidden state 
$\mathbf{h}_t \in \mathbb{R}^d$ is mapped to a sparse feature vector 
$\mathbf{z}_t \in \mathbb{R}^n$ with $n \gg d$, via
\[
\mathbf{z}_t =
\mathrm{TopK}\!\left(
\mathrm{ReLU}(\mathbf{W}_{\mathrm{enc}}\mathbf{h}_t+\mathbf{b}), k
\right),
\qquad
\hat{\mathbf{h}}_t=\mathbf{z}_t\mathbf{W}_{\mathrm{dec}}^\top .
\]
Because the SAE decoder is linear in $\mathbf{z}_t$, differences in feature
space map directly to differences in hidden space:
$\Delta \mathbf{h}_t = \Delta \mathbf{z}_t \mathbf{W}_{\mathrm{dec}}^\top$.
We exploit this property by defining the intervention in the sparse SAE
space, where target-aligned features can be identified and edited, and then
applying the corresponding decoded update directly in hidden space.


\subsection{Target-concept Direction}
\label{sec:ranking}

Given a target concept $c$, we partition the full prompt corpus into a 
\emph{forget set} $\mathcal{P}_f = \{p : c \in p\}$ of prompts mentioning the 
target and a \emph{retain set} $\mathcal{P}_r = \{p : c \notin p\}$ of all 
remaining prompts.
Each prompt is tokenised into up to a fixed-length sequence of $T$.
Usually, most token positions 
encode generic content (articles, prepositions, style descriptors) with a small 
number encoding the target identity itself (e.g.\ the tokens \textit{barack} 
and \textit{obama} in ``A realistic portrait of Barack Obama''). 

Because all 
prompts in $\mathcal{P}_f$ share the target name and tokenise the 
target concept at similar positions, 
SAE features associated with the target identity are expected to activate
systematically at those positions across $\mathcal{P}_f$ and remain less active
at the same positions across $\mathcal{P}_r$ (where the 
prompt mentions different identities).
The forget-retain contrast at each 
token position therefore isolates identity-specific features from generic 
content features. Figure~\ref{fig:alpha_heatmap_app} in the supplementary appendix confirms this empirically: the highest-scoring features concentrate their activation at the name-bearing token positions.

Let $\mathcal{V} \subseteq \{1,\ldots,T\}$ denote the set of non-padding
token positions for a given prompt. For each feature index $i$ and
position $t \in \mathcal{V}$, we compute the means and standard
deviations of $z_{ti}$ across the forget and retain sets:
\begin{equation}
\mu^f_{ti} = \mathbb{E}_{p \in \mathcal{P}_f}\!\bigl[\mathbf{z}_t(p)[i]\bigr],
\quad
\sigma^f_{ti} = \mathrm{Std}_{p \in \mathcal{P}_f}\!\bigl[z_{ti}(p)\bigr],
\end{equation}
and analogously $\mu^r_{ti}, \sigma^r_{ti}$ over $\mathcal{P}_r$.

We then score each feature by its discriminative power, averaged over valid token 
positions:
\begin{equation}
\mathcal{S}_i =
\frac{
\tfrac{1}{|\mathcal{V}|} \sum_{t \in \mathcal{V}} (\mu^f_{ti} - \lambda_r \mu^r_{ti})
}{
\tfrac{1}{|\mathcal{V}|} \sum_{t \in \mathcal{V}} (\sigma^f_{ti} + \sigma^r_{ti}) + \varepsilon
}.
\end{equation}
The numerator favours features that fire strongly across the forget set and 
weakly across the retain set, modulated by a retain penalty $\lambda_r$. 
The 
denominator penalises features whose retain-set activations vary unpredictably 
from prompt to prompt: a feature with high $\sigma^r$ does not have a stable 
``baseline'' activation on non-target prompts, so subtracting it would induce 
inconsistent edits across non-target generations. 
We select the top-$k$ 
features and 
represent the resulting index set with $\mathcal{I}^*$. 
Ablations 
over $k$ and $\lambda_r$ are reported in Section~\ref{sec:abl}.

From the top-$k$ features we form a per-token direction in SAE space pointing 
from the retain centroid to the forget centroid:
\begin{equation}
\mathbf{d}_t = \boldsymbol{\mu}^f_t - \boldsymbol{\mu}^r_t \in \mathbb{R}^n
\;\;(\text{coordinates outside } \mathcal{I}^* \text{ set to zero}),
\qquad
\hat{\mathbf{d}}_t = \frac{\mathbf{d}_t}{\|\mathbf{d}_t\|_2}.
\end{equation}
The unit vector $\hat{\mathbf{d}}_t$ encodes the feature-space deviation of 
the target concept from the retain distribution at token position $t$.

\subsection{Counterfactual Projection}
\label{sec:hook}

At generation time, we apply a forward hook at the layer-$\ell$ output of the 
conditional branch of the text encoder. For each token position $t$, let 
$\mathbf{h}_t$ be the dense hidden representation and 
$\mathbf{z}_t =
\mathrm{TopK}\!\left(
\mathrm{ReLU}(\mathbf{W}_{\mathrm{enc}}\mathbf{h}_t+\mathbf{b}), k
\right)$
its k-SAE encoding. The hook performs three operations.

\paragraph{Step 1: Measure the alignment signal.}
We compute a scalar that captures how much of the concept direction is present 
in the sparse encoding at token position $t$:
\begin{equation}
\alpha_t = \max\!\bigl((\mathbf{z}_t - \boldsymbol{\mu}^r_t) \cdot \hat{\mathbf{d}}_t,\; 0\bigr) \;\geq\; 0.
\label{eq:alpha}
\end{equation}
Here $\hat{\mathbf{d}}_t$ is the unit target-concept axis: the dot product projects $\mathbf{z}_t - \boldsymbol{\mu}^r_t$ onto this axis to measure the target component.
$\alpha_t$ is large when the prompt strongly 
invokes the target at position $t$ and near zero otherwise. The clamp 
prevents amplifying the concept signal when the 
sparse
activation is already 
towards the retain side.

\paragraph{Step 2: Projection in SAE space.}
We then define a counterfactual SAE code by removing an $s\alpha_t$-scaled 
component along the same direction:
\begin{equation}
\mathbf{z}_t^{\mathrm{cf}} = \mathbf{z}_t - s\,\alpha_t\,\hat{\mathbf{d}}_t,
\qquad s \in [0,1].
\label{eq:proj}
\end{equation}
The same target-concept axis $\hat{\mathbf{d}}_t$ is used twice:
Eq.~\eqref{eq:alpha} measures the target component, and
Eq.~\eqref{eq:proj} subtracts that measured component along the same axis. 
The scalar $s$ controls the strength of the counterfactual intervention: $s=0$ recovers the original representation, while larger values of $s$ produce increasingly counterfactual representations by shifting the SAE features away from the target-concept direction. At $s=1$, this is the Euclidean projection onto the \emph{closed retain half-space}
$\mathcal H^-_t=\{u:(u-\boldsymbol{\mu}^r_t)^\top\hat{\mathbf d}_t\leq0\}$:
tokens already on the retain side receive no edit, while positively aligned
tokens are projected onto its boundary. For intermediate strengths, the update
interpolates between the original code and this projection.


\begin{proposition}[Bounded counterfactual intervention]
Let $q_t=(\mathbf z_t-\boldsymbol\mu^r_t)^\top\hat{\mathbf d}_t$ and
$\alpha_t=[q_t]_+=\max(q_t,0)$. For $s\in[0,1]$, the update
$\mathbf z_t^{\mathrm{cf}}=\mathbf z_t-s\alpha_t\hat{\mathbf d}_t$
satisfies
\[
(\mathbf z_t^{\mathrm{cf}}-\boldsymbol\mu^r_t)^\top\hat{\mathbf d}_t
= q_t-s[q_t]_+ =
\begin{cases}(1-s)q_t,&q_t>0,\\q_t,&q_t\leq0.\end{cases}
\]
Thus positive target-direction evidence decreases without crossing the
retain-side boundary, while nonpositive evidence is left unchanged. At
$s=1$, $\mathbf z_t^{\mathrm{cf}}=\Pi_{\mathcal H^-_t}(\mathbf z_t)$,
the unique minimum-Euclidean-distance projection of $\mathbf z_t$ onto
$\mathcal H^-_t$ in SAE coordinates.
\end{proposition}
Full proof is available in Supplementary Appendix~\ref{app:proofs}

\paragraph{Step 3: Decode to hidden space.}
Because the SAE decoder is linear, the SAE-space change
$\mathbf{z}_t - \mathbf{z}_t^{\mathrm{cf}} = s\alpha_t\hat{\mathbf{d}}_t$
maps directly to a corresponding hidden-space update. Thus, rather than
decoding the full counterfactual code, we can apply only the decoded
intervention direction to the original hidden state:
\begin{equation}
\mathbf{h}^{\mathrm{CFS}}_t = \mathbf{h}_t \;-\; s\,\alpha_t\,\bigl(\hat{\mathbf{d}}_t\,\mathbf{W}_{\mathrm{dec}}^{\!\top}\bigr).
\label{eq:steered}
\end{equation}
Here, $\hat{\mathbf{d}}_t \mathbf{W}_{\mathrm{dec}}^{\!\top} \in \mathbb{R}^d$
is the hidden-space direction induced by the target-concept direction in
the SAE latent space. 
These decoded directions are precomputed once per target, avoiding the need
to explicitly construct the counterfactual code
$\mathbf{z}_t^{\mathrm{cf}}$ at inference time.

Taken together, these steps implement CFS as a lightweight inference-time
intervention: the target direction is identified once in SAE space, decoded
once into hidden space, and then applied per token according to the live
alignment score $\alpha_t$. As a result, prompts that do not activate the
target direction receive little or no update, while prompts containing the
target concept are steered toward a counterfactual representation with the
target-aligned component suppressed.

\section{Experiments \& Results}
\label{sec:experiments}

\subsection{Experimental setup}
\label{sec:setup}

\paragraph{Model and baselines.}
We evaluate on Stable Diffusion~1.4~\citep{rombach2022high} and
compare CFS against six baselines spanning both inference-time
steering and weight-editing approaches: Concept
Steerers~\citep{kim2025concept} (SAE-based fixed-vector
subtraction), SAeUron~\citep{cywinski2025saeuron} (SAE feature
suppression on the UNet), One-Step~\citep{surkov2024one} (fixed
mean concept vector), SLD~\citep{schramowski2023safe} (inference-time
CFG suppression), ESD~\citep{gandikota2023erasing} (UNet fine-tuning),
and UCE~\citep{gandikota2024unified} (closed-form weight editing).
\paragraph{Baseline evaluation protocol.}
All methods use the same identity targets, prompts, seeds, inference steps, and guidance settings in the reported comparison. CFS is evaluated at full steering strength $s=1$; strength-controlled baselines are reported at their best AWUS operating point over the evaluated strength sweeps, while ESD and UCE use the corresponding edited checkpoints. Fixed-full-strength CLIP results appear in Supplementary Appendix~\ref{app:fixed_strength}. Complete implementation and evaluation details are documented in Supplementary Appendix~\ref{app:reproducibility}.

\paragraph{Identity unlearning targets.}
To our knowledge, no standard benchmark currently exists for identity unlearning while preserving adjacent identities. In this work, we study this setting through adjacency preservation, which requires maintaining each target identity’s associations with related identities after unlearning.
To evaluate this problem, we construct a corpus of 4,000 prompts derived from Celeb-100, covering 100 celebrity identities across politics, sports, entertainment, and technology. 
From this corpus, we select eight target identities spanning diverse domains: Barack Obama, Joe Biden, Elon Musk, Bill Gates, Lionel Messi, Cristiano Ronaldo, Bradley Cooper, and Scarlett Johansson. 
For each target we define the forget set $\mathcal{P}_f$
as all prompts naming that identity, and the retain set $\mathcal{P}_r$
as all remaining prompts. 
We evaluate adjacency preservation using 4 adjacent identities per target,
chosen based on semantic relatedness and image-space similarity
(Supplementary Appendix~\ref{app:identity_adjacency}; Figure~\ref{fig:identity_similarity};
full list in Supplementary Appendix~\ref{app:targets}), and 9 generic prompts.

\paragraph{Model-internal adjacency check.}
As a complementary analysis to semantic and CLIP-image-based peer selection, we average name-token representations from SD~1.4's frozen layer-9 text encoder across prompts for 100 identities and compute pairwise cosine similarity. This analysis uses raw model conditioning representations rather than SAE features or generated-image evaluation scores. Examples include Ronaldo--Messi (0.460), Obama--Biden (0.394), Musk--Zuckerberg (0.401), Cooper--Gosling (0.421), and Johansson--Portman (0.506). For the 13 manually selected peers present in this identity corpus, the ranking achieves MRR 0.487. This establishes a model-grounded view of proximity, not a formal or unique definition of adjacency. Layer~9 was selected by the downstream intervention ablation, independently of this post-hoc adjacency analysis. Additional examples appear in Supplementary Appendix~\ref{app:model_internal_adj}.


\paragraph{Generalization experiments.}
Beyond identity, we further evaluate CFS on nudity unlearning using the I2P benchmark~\citep{schramowski2023safe} and on style unlearning using prompts targeting Van Gogh. For nudity, we use an SAE trained on I2P prompts; for style, we use an SAE trained on the style-prompt corpus from Unlearn Canvas~\citep{zhang2024unlearncanvas} and apply it at layer~9. We use the same CFS hyperparameters across all experiments, $k=48$ and $\lambda_r=0.1$, without changing the projection rule. These experiments test cross-domain transfer on SD 1.4, which confirms the method's generalization capabilities.

\subsection{Evaluation metrics}
\label{sec:metrics}
To quantify whether unlearning a target identity also erases information about adjacent identities, we introduce the \textbf{Adjacency-Weighted Unlearning Score (AWUS)}. At a high level, AWUS captures the tradeoff
\[
\mathrm{AWUS} = \text{Target Erasure} - \text{Collateral Damage},
\]
where the first term measures how effectively the target identity is erased, and the second penalizes unintended changes to adjacent and generic prompts. AWUS is computed in CLIP embedding space, so adjacency reflects semantic similarity between identities.

To capture identity-specific preservation beyond semantic similarity, we extend AWUS to a face-focused embedding space using ArcFace, yielding \textbf{FaceAWUS}. FaceAWUS uses the same target-erasure versus collateral-damage structure, but defines adjacency using facial identity similarity rather than CLIP-space semantic similarity.
We further measure collateral forgetting using an \textbf{LPIPS-gap} metric, which quantifies how much more the target image changes than adjacent and generic images after unlearning. Together, AWUS, FaceAWUS, and LPIPS-gap provide complementary views across semantic, face-identity, and perceptual spaces: the target concept should change substantially, while adjacent and generic prompts should remain stable. 
CLIP-based scores and whole-image LPIPS can respond to backgrounds and other attributes as well as identity; ArcFace evaluation of detected faces provides complementary identity-specific evidence, but depends on the recognizability of the baseline face and on successful detection.
Formal definitions are provided in Supplementary Appendix~\ref{app:eval_metrics}.

\subsection{Results}
\label{sec:main}

\textbf{Main Results.} Table~\ref{tab:main} reports the three net utility metrics averaged
over 8 celebrity targets. CFS achieves the highest AWUS ($0.296$),
FaceAWUS ($0.636$), and LPIPS-gap ($0.402$) across all methods. 
Methods with higher raw erasure scores achieve them at $12$--$22\times$ the collateral damage, which the net metrics correctly penalise.

\begin{table}[t]
\centering
\caption{%
  \textbf{Identity unlearning: net utility metrics.}
  We report the three metrics defined in Section~\ref{sec:metrics};
  each rewards target erasure while penalising adjacency damage. Higher is better for all metrics.
  \textbf{Bold}: best. Per-target results are reported in
  Supplementary Appendix~\ref{app:per_target}.
}
\label{tab:main}
\small
\setlength{\tabcolsep}{6pt}
\begin{tabular}{lccc}
\toprule
Method & AWUS\,$\uparrow$ & FaceAWUS\,$\uparrow$ & LPIPS-gap\,$\uparrow$ \\
\midrule
SAeUron          & 0.233 &    0.300 & 0.108 \\
Concept Steerers & 0.279 & $-$0.077 & 0.038 \\
UCE              & 0.198 &    0.543 & 0.163 \\
ESD              & 0.262 &    0.138 & 0.051 \\
SLD              & 0.003 &    0.093 & 0.022 \\
One-Step         & 0.106 &    0.136 & 0.033 \\
\midrule
\textbf{CFS (Ours)} & \textbf{0.296} & \textbf{0.636} & \textbf{0.402} \\
\bottomrule
\end{tabular}
\end{table}

\paragraph{Independent face-identity validation.}
To separate face-identity changes from changes to background and composition, we evaluate detected face regions with ArcFace under paired prompt/seed/generation settings. References are constructed from unmodified SD~1.4 generations, not from real celebrity photographs. At full CFS strength, mean target-face similarity is 0.160 ($n=38$), while adjacent identities retain similarity 0.760 ($n=56$). Face detection fails for 20.8\% of target outputs and 3.4\% of peer outputs. These results are consistent with greater target-face disruption than peer-face disruption, but detection failures and variability of baseline identity rendering constrain interpretation. The unmodified model's self-consistency across three prompt templates ranges from approximately 0.15 to 0.66; most reported targets fall around 0.45--0.66. Lower-consistency targets are weaker tests of meaningful identity removal.

\paragraph{Preservation of general generation quality.}
We evaluate 10,000 COCO-30K captions at full CFS strength, averaging over all eight identity directions. Baseline and edited generations use identical captions, seeds, sampler, guidance, and inference steps; FID is calculated against the same corresponding real COCO images. Unmodified SD~1.4 has FID 15.71 and CLIPScore 27.39, versus FID 16.11 and CLIPScore 26.95 for CFS (changes of +0.40 and $-0.44$, respectively). This evaluates broad generation preservation beyond the selected identities.

\begin{figure*}[t]
\centering
\includegraphics[scale=0.75]{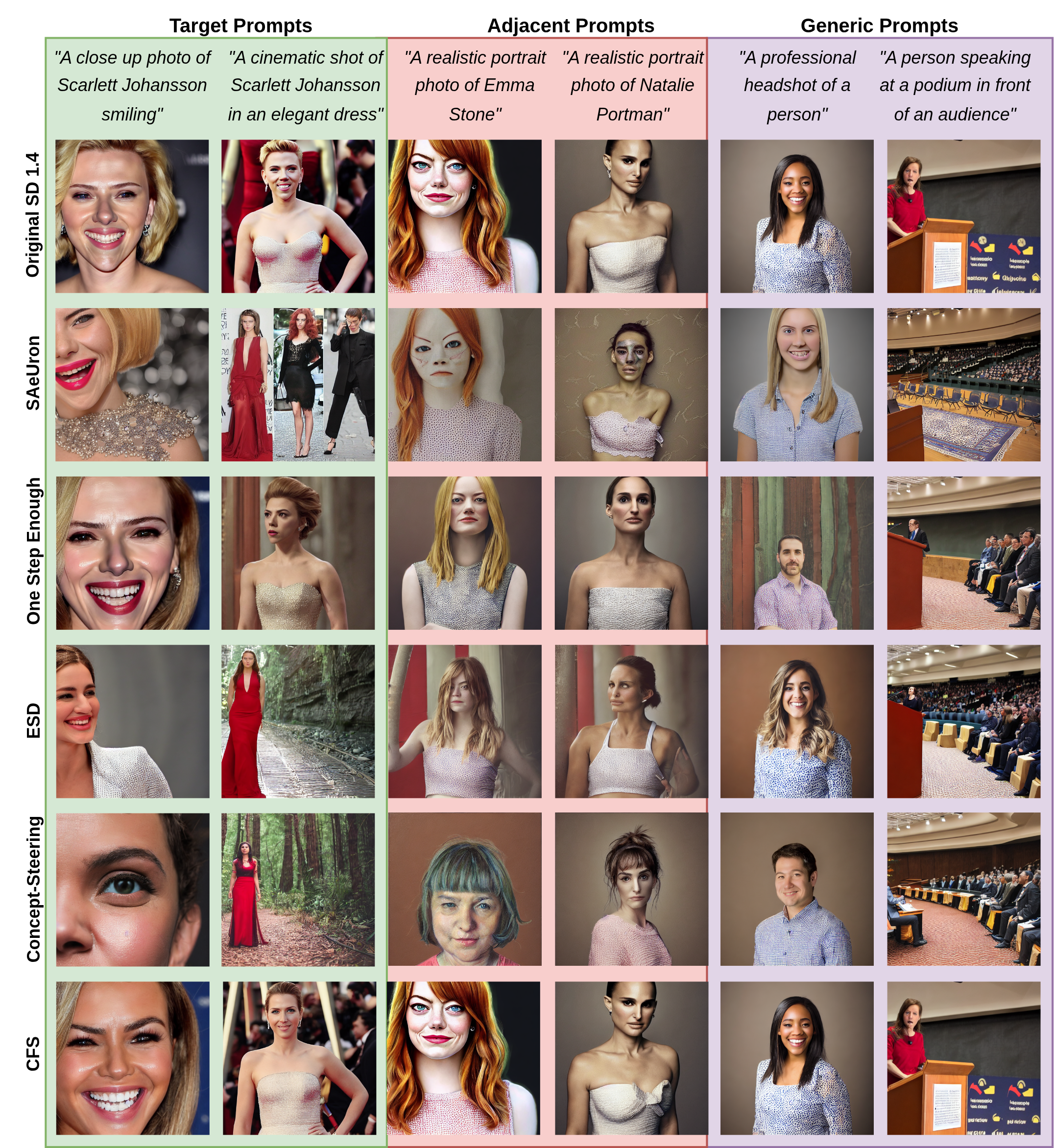}

\caption{%
\textbf{Adjacency preservation under identity concept removal.}
CFS removes the target identity while preserving adjacent and generic prompts.
}
\vspace{-1em}
\label{fig:adjacency_comparison}
\end{figure*}

Figure~\ref{fig:adjacency_comparison} provides a side-by-side qualitative
comparison of target removal and collateral damage. The green columns show
target prompts generation for the erased identity, Scarlett Johansson; the pink columns
show adjacent identities, Emma Stone and Natalie Portman; and the light blue columns
show generic person prompts generation after the identity removal. CFS removes the target identity while preserving adjacent and generic identities along with the scene attributes. In contrast, prior methods with high collateral damage introduce noticeable distortions in adjacent and generic identities, reflecting their lack of input adaptivity and feature disentanglement.

\begin{figure*}[!t]
\centering
\includegraphics[width=\textwidth]{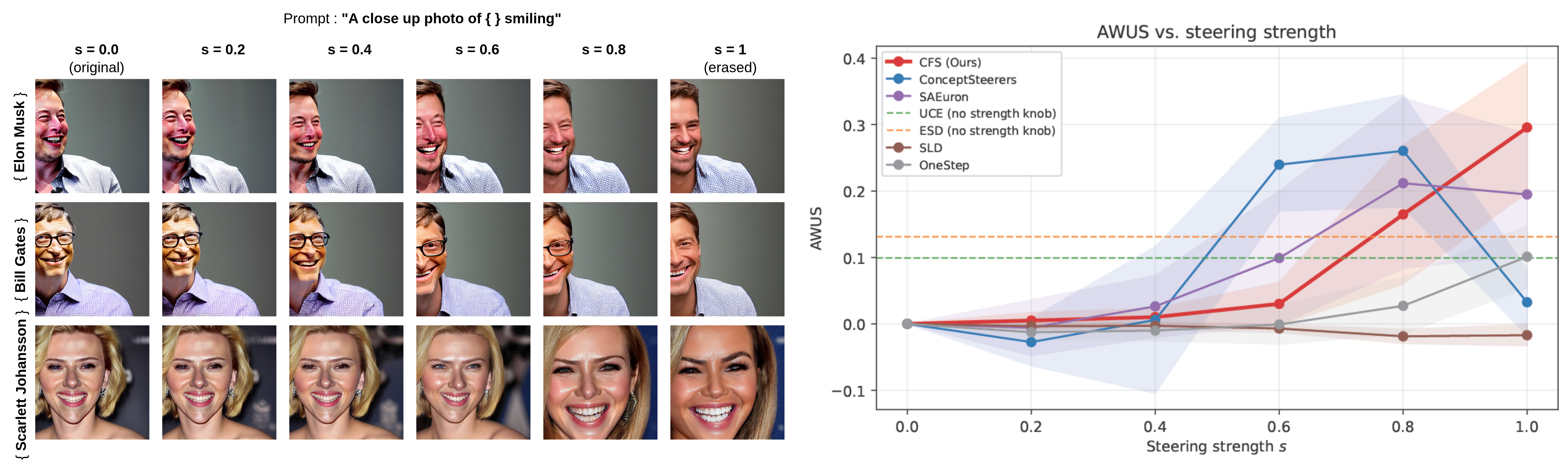}
\caption{%
\textbf{Strength stability.}
Left: qualitative sweep over steering strength $s$.
Right: AWUS across strengths; dashed lines denote weight-editing methods
without a test-time strength parameter.
}
\label{fig:strength_stability}
\end{figure*}

\textbf{Strength stability.} Figure~\ref{fig:strength_stability} shows that CFS remains stable across
projection strengths. Qualitatively, identity is progressively removed as
$s$ increases while pose, lighting, and background remain stable. Quantitatively,
AWUS improves up to full projection, whereas subtraction-based methods often
degrade at high strength due to unbounded edits. A face-space trade-off plot in Supplementary Appendix~\ref{app:face_tradeoff} further
shows that CFS achieves high target-identity change with low adjacent-identity drift.


\begin{figure*}[t]
\centering
\includegraphics[width=\textwidth]{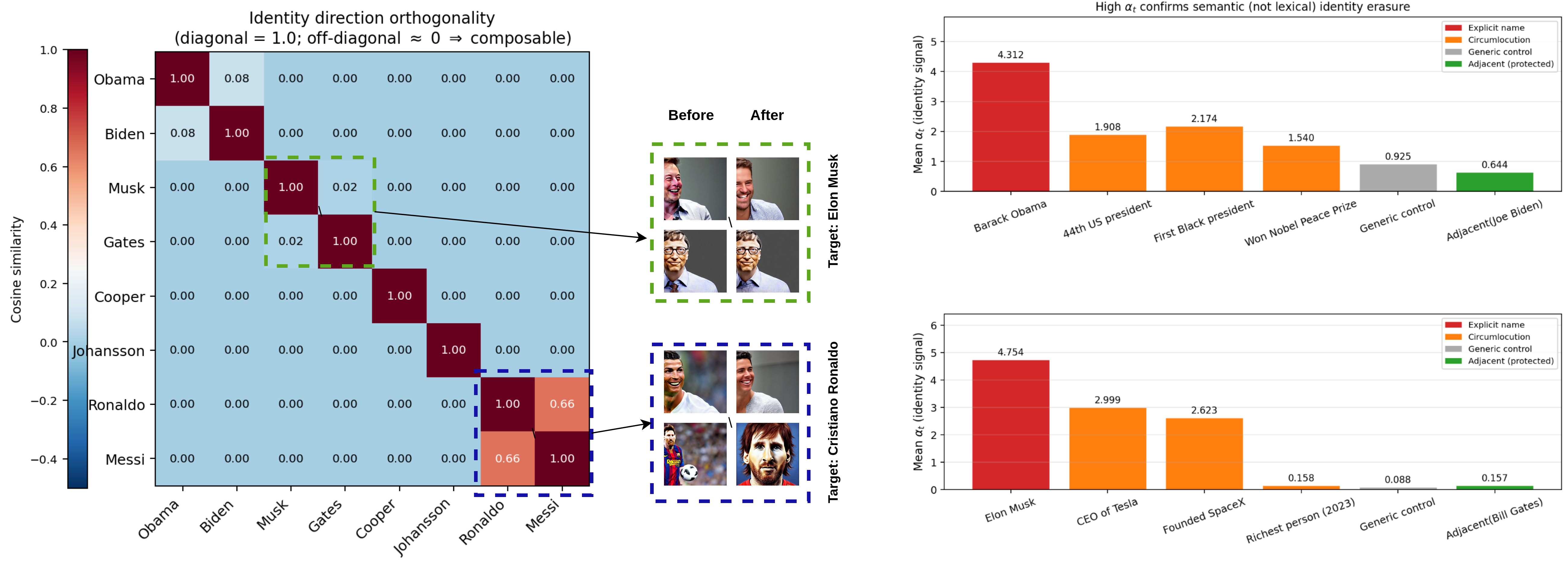}
\caption{%
\textbf{Identity-direction geometry and semantic alignment.}
Left: pairwise cosine similarity between CFS identity directions.
Right: alignment signals for explicit names, semantic circumlocutions,
generic controls, and adjacent identities.
}
\vspace{-1em}
\label{fig:geometry_semantic}
\end{figure*}

\textbf{Identity geometry and semantic alignment.} Figure~\ref{fig:geometry_semantic} analyzes why CFS preserves adjacent
identities. Most measured identity directions have low pairwise cosine similarity, consistent with limited overlap along the chosen axes, but orthogonality is not guaranteed; Ronaldo--Messi illustrates a more entangled pair for which collateral effects remain possible. The alignment signal (Figure~\ref{fig:geometry_semantic}, right) responds to semantic descriptions: for Elon Musk, ``CEO of Tesla''
and ``Founded SpaceX'' reach $63\%$ and $55\%$ of the explicit-name signal,
while Bill Gates remains low (bottom). Similarly, Obama-related circumlocutions
activate the target direction more strongly than the adjacent identity
Joe Biden (top panel). Together, these results show that CFS captures semantic identity
signals while limiting activation on adjacent identities. A token-level heatmap in Supplementary Appendix~\ref{app:token_alignment} confirms that
the alignment signal concentrates on identity-bearing tokens.

Additional qualitative results in Supplementary Appendix~\ref{app:compositional} show
that CFS can compose two identity directions in a single generation and
can erase identities from circumlocution prompts.

\textbf{Generalisation: Nudity and Style unlearning.} Although our primary focus is identity unlearning, CFS is concept-agnostic:
the same projection mechanism can be applied to other concept unlearning tasks. We test this on two additional settings: nudity suppression and artistic-style removal.

\begin{figure*}[!t]
\centering
\includegraphics[width=\textwidth]{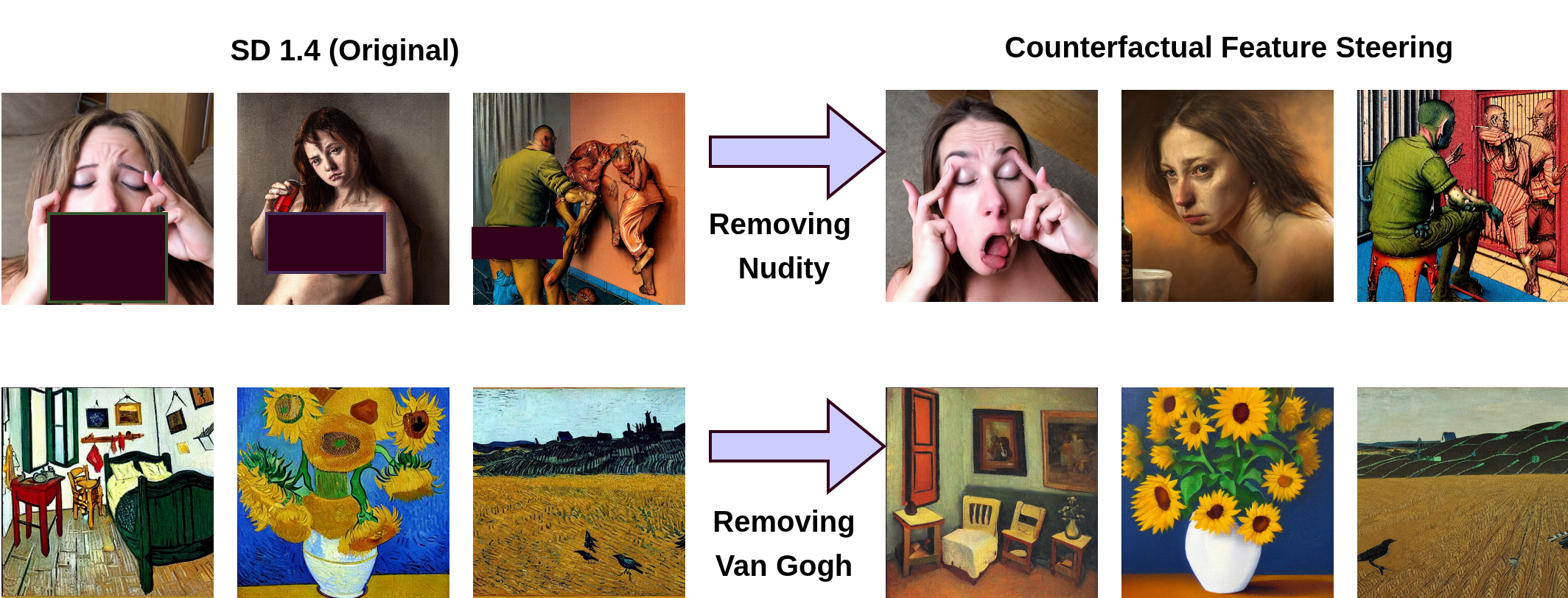}
\caption{%
  \textbf{Generalisation beyond identity unlearning.}
  Top: nudity suppression. Bottom: Van Gogh-style removal. CFS reduces
  the target concept while preserving coarse scene structure.
}

\label{fig:generalisation}
\end{figure*}

Figure~\ref{fig:generalisation} shows that the same CFS mechanism extends
beyond identity removal. In the unsafe-content examples, the intervention
reduces the targeted visual content while largely preserving the coarse scene
layout. In the style examples, CFS progressively removes recognizable Van
Gogh-style features while maintaining the object and scene structure, such as
the bedroom layout, sunflower arrangement, and wheat-field composition. These
results suggest that CFS is not restricted to identity-specific directions;
rather, it can operate on different concept directions learned in SAE feature
space.

\paragraph{Quantitative nudity-suppression transfer.}
Table~\ref{tab:nudity_transfer} reports Nudity Detection Rate (\%, lower is better) from the rebuttal experiments on I2P, Ring-a-Bell, and MMA-Diffusion. CFS outperforms inference-time SLD-MAX in all reported settings and gives the lowest rates on I2P and Ring-a-Bell; ESD has the lowest rate on MMA-Diffusion. These results support transfer to a safety-related concept on SD~1.4.
\begin{table}[t]
\centering
\caption{\textbf{Nudity suppression across standard and adversarial prompt sets.} Detection rate (\%, lower is better); SD~1.4 is unmodified.}
\label{tab:nudity_transfer}
\small\setlength{\tabcolsep}{5pt}
\begin{tabular}{lrrrr}
\toprule
Benchmark & SD~1.4 & ESD & SLD-MAX & CFS \\
\midrule
I2P & 17.80 & 2.87 & 1.74 & \textbf{0.47} \\
Ring-a-Bell K77 & 85.26 & 20.00 & 23.16 & \textbf{4.27} \\
Ring-a-Bell K38 & 87.37 & 29.47 & 32.63 & \textbf{7.42} \\
Ring-a-Bell K16 & 93.68 & 35.79 & 42.11 & \textbf{12.63} \\
MMA-Diffusion & 95.70 & \textbf{12.70} & 52.10 & 17.48 \\
\bottomrule
\end{tabular}
\end{table}

\vspace{-1em}

%

\subsection{Ablation}
\label{sec:abl}
We ablate the main CFS hyperparameters in a single-target setting using the identity ``Elon Musk'' to understand how feature sparsity, retain strength, and intervention layer affect unlearning utility. Table~\ref{tab:ablations_combined} shows that larger values of $k$ generally provide a stronger signal, while performance remains relatively stable across values of $\lambda_r$. The layer ablation further shows that intervention effectiveness varies across layers, with middle layers performing best, consistent with prior mechanistic interpretability findings on selecting effective layers for SAE training \citep{templeton2024scaling}. We also analyze the computational cost of CFS in Supplementary Appendix~\ref{app:runtime}; the reported per-image time is close to unmodified SD~1.4, with domain-level SAE training separately accounted for.

\vspace{-1em}

\begin{table*}[t]
\centering
\caption{
\textbf{Ablation studies.}
\textbf{(Left)} Hyperparameter ablation reports AWUS at $s=1$ for each $(k,\lambda_r)$ pair in a single-target setting.
\textbf{(Right)} Layer ablation reports AWUS across intervention layers.
}
\label{tab:ablations_combined}

\small
\setlength{\tabcolsep}{4.2pt}

\begin{minipage}[t]{0.72\textwidth}
\centering

\begin{tabular}{lcccccc}
\toprule
\multirow{2}{*}{\textbf{top-}$k$}
  & \multicolumn{6}{c}{\textbf{Retain weight} $\lambda_r$} \\
\cmidrule(lr){2-7}
  & 0.1 & 0.2 & 0.4 & 0.6 & 0.8 & 1.0 \\
\midrule
5   & 0.000 & 0.000 & 0.000 & 0.000 & 0.000 & 0.028 \\
10  & 0.000 & 0.000 & 0.000 & 0.000 & 0.000 & 0.003 \\
20  & 0.000 & 0.000 & 0.000 & 0.000 & 0.000 & 0.360 \\
30  & 0.001 & 0.001 & 0.001 & 0.001 & 0.001 & 0.366 \\
48  & \textbf{0.375} & 0.374 & 0.375 & 0.350 & 0.350 & 0.331 \\
64  & 0.347 & 0.347 & 0.355 & 0.339 & 0.327 & 0.341 \\
96  & 0.317 & 0.308 & 0.323 & 0.353 & 0.366 & 0.303 \\
100 & 0.327 & 0.327 & 0.326 & 0.347 & 0.342 & 0.305 \\
150 & 0.310 & 0.339 & 0.305 & 0.302 & 0.351 & 0.335 \\
200 & 0.328 & 0.307 & 0.317 & 0.334 & 0.313 & 0.293 \\
\bottomrule
\end{tabular}
\end{minipage}
\hfill
\begin{minipage}[t]{0.23\textwidth}
\centering

\begin{tabular}{lc}
\toprule
Layer $\ell$ & AWUS\,$\uparrow$ \\
\midrule
4            & 0.167 \\
6            & 0.245 \\
8            & 0.311 \\
9            & \textbf{0.375} \\
10           & 0.302 \\
11           & 0.288 \\
\bottomrule
\end{tabular}
\end{minipage}

\end{table*}

\paragraph{Necessity of the SAE representation.}
We compare the same bounded, input-adaptive projection in layer-9 SAE space with a projection applied directly to the raw layer-9 hidden states, keeping prompts, target identities, seeds, and strengths fixed. The raw-space method achieves slightly higher target erasure and AWUS, but SAE-space CFS has lower CLIP collateral drift and higher peer-face similarity (Table~\ref{tab:sae_necessity}). Thus, the evidence supports the SAE's contribution to selectivity rather than claiming it always improves every scalar metric.
\begin{table}[t]
\centering
\caption{\textbf{SAE-space versus raw hidden-space projection.} Under matched experimental settings; this separate ablation should not be numerically conflated with the eight-target averages in Table~\ref{tab:main}.}
\label{tab:sae_necessity}
\small
\begin{tabular}{lcc}
\toprule
Metric & SAE-space CFS & Raw hidden-space \\
\midrule
Target erasure rate $\uparrow$ & 0.387 & \textbf{0.399} \\
CLIP collateral drift $\downarrow$ & \textbf{0.0065} & 0.0078 \\
AWUS $\uparrow$ & 0.381 & \textbf{0.391} \\
Peer face similarity retained $\uparrow$ & \textbf{0.848} & 0.722 \\
\bottomrule
\end{tabular}
\end{table}


\FloatBarrier
\section{Conclusion}

In this work, we study the challenge of adjacency preservation within identity unlearning. To address this problem, we introduce Counterfactual Feature Steering (CFS), an inference-time identity-suppression method requiring one-time domain-level SAE training that identifies a direction in sparse feature space capturing the target concept while avoiding features associated with retained concepts. CFS then applies per-token projection at inference time, editing sparse features along this direction to produce counterfactual representations. To quantify whether unlearning a target identity also erases information about adjacent identities, we introduce the Adjacency-Weighted Unlearning Score (AWUS) and its extensions across complementary representation spaces.
Using a corpus of 4,000 prompts derived from Celeb-100, covering 100 celebrity identities across politics, sports, entertainment, and technology, we find that CFS achieves the strongest tradeoff between target erasure and preservation for selected target identities. Moreover, CFS remains stable across steering strengths, whereas subtraction-based methods collapse under stronger interventions. We further show that the same projection rule yields quantitative and qualitative nudity and artistic-style suppression results.

Despite the growing importance of identity unlearning and the need to preserve non-target identities, including adjacent identities, a key limitation of this 
line of research
is the lack of a systematic benchmark for evaluating this setting. While we make a best-effort attempt to curate adjacent identities using public resources such as Celeb-100, and further evaluate both semantic relatedness and image-space similarity, developing a dedicated benchmark for identity unlearning with adjacency preservation remains an important direction for future work. More broadly, CFS and related concept-steering methods depend on the quality of the underlying SAE representation. A deeper analysis of SAE learning and its ability to capture concept-specific features is therefore another important direction for future research.

\clearpage

\bibliographystyle{plainnat}
\bibliography{main}


\clearpage
\ifdefined\CFSEmbedSupplement
  \appendix
\numberwithin{figure}{section}
\numberwithin{table}{section}

\ifdefined\CFSStandaloneSupplement
\else
\begin{center}
{\LARGE \textbf{Appendix}}
\end{center}
\fi

\section{Target and Adjacent Identity Configuration}
\label{app:targets}

Table~\ref{tab:targets} lists the 8 target identities, their Adjacent
identities used for adjacency evaluation, and the 9 shared generic
prompts evaluated for all targets.

\begin{table}[h]
\centering
\caption{Target identities and their adjacent sets.}
\label{tab:targets}
\small
\setlength{\tabcolsep}{4pt}
\begin{tabular}{ll}
\toprule
\textbf{Target} & \textbf{Adjacent identities} \\
\midrule
Barack Obama    & Biden, Trump, Modi, Lincoln \\
Joe Biden       & Obama, Trump, Modi, Lincoln \\
Elon Musk       & Bezos, Zuckerberg, Gates, Cook \\
Bill Gates      & Musk, Bezos, Zuckerberg, Cook \\
Bradley Cooper  & Reynolds, Gosling, Hemsworth, Jackman \\
Scarlett Johansson & Hathaway, Stone, Portman, Kunis \\
Cristiano Ronaldo & Messi, Neymar, Modrić, Ramos \\
Lionel Messi    & Ronaldo, Neymar, Modrić, Ramos \\
\midrule
\textbf{Generic prompts (shared)} & businessman, man in black suit, \\
& professional headshot, conference speaker, \\
& podium speaker, SpaceX rocket launch, \\
& formal portrait, person playing football, \\
& person on city street \\
\bottomrule
\end{tabular}
\end{table}

\section{Proof of Proposition 1}
\label{app:proofs}

We restate proposition 1 mentioned in Section~\ref{sec:method} with full proof. We use the notation of
Section~\ref{sec:method}: $\zvec_t\in\R^n$ is the SAE encoding,
$\mur\in\R^n$ the per-token retain centroid restricted to $\mathcal{I}^*$,
$\dhat\in\R^n$ the unit concept direction restricted to $\mathcal{I}^*$
(with $\|\dhat\|_2=1$), $\alpha_t=\max((\zvec_t-\mur)\cdot\dhat,0)\geq0$,
and $\zvec^{\mathrm{cf}}_t=\zvec_t-s\alpha_t\dhat$ for $s\in[0,1]$.


\begin{proof}
Write $q_t=(\zvec_t-\mur)^\top\dhat$. Since $\|\dhat\|_2=1$ and
$\alpha_t=[q_t]_+$, substitution yields
\[
(\zvec_t^{\mathrm{cf}}-\mur)^\top\dhat
=q_t-s\alpha_t=q_t-s[q_t]_+.
\]
If $q_t>0$, this equals $(1-s)q_t\geq0$; in particular, $s=1$
brings the positive component to zero without crossing the boundary.
If $q_t\leq0$, then $\alpha_t=0$ and the code is unchanged, with its
nonpositive signed component preserved. These cases prove the first claim.
For $s=1$, the closed half-space
$\mathcal H^-_t=\{u:(u-\mur)^\top\dhat\leq0\}$ has Euclidean
projection $\Pi_{\mathcal H^-_t}(\zvec_t)=\zvec_t-[q_t]_+\dhat$.
This is the standard orthogonal projection onto a closed convex half-space,
and is the unique Euclidean-distance minimizer among points in
$\mathcal H^-_t$. For $s\in[0,1]$, CFS interpolates between
$\zvec_t$ and that projection.
\end{proof}

\section{Evaluation metrics definition}
\label{app:eval_metrics}

For each configuration (method, target, strength), we generate images
for the target prompts, adjacent prompts, and generic prompts. Let $g_p^0$
and $g_p^s$ denote images generated from prompt $p$ by the unmodified
model and the steered model at strength $s$, respectively. We denote
the set of target prompts by $\mathcal{T}$, adjacent prompts by
$\mathcal{A}$, generic prompts by $\mathcal{G}$, and all non-target
prompts by $\mathcal{N}=\mathcal{A}\cup\mathcal{G}$.
\paragraph{CLIP-space metrics.}
Let $\mathrm{CLIP}(p,s)=\cos(\psi(g_p^s),\psi_t(p))$ be the cosine
similarity between the CLIP image embedding and the CLIP text
embedding of prompt $p$. Define the normalised score drop:
\[
  \delta(p,s) = \max\!\left(0,\;\frac{\mathrm{CLIP}(p,0)-\mathrm{CLIP}(p,s)}{|\mathrm{CLIP}(p,0)|+\varepsilon}\right).
\]

With this, we define the \textbf{Adjacency-Weighted Unlearning Score (AWUS)}:
\[
  \mathrm{AWUS}(s)
  =
  \mathbb{E}_{p\in\mathcal{T}}[\delta(p,s)]
  -
  \mathbb{E}_{p\in\mathcal{N}}[\delta(p,s)].
\]
where the first term measures target erasure and the second measures
collateral damage to adjacent and generic prompts. Higher AWUS is better.
This formulation rewards target removal while penalising methods that
achieve high erasure by damaging the surrounding representation manifold.

\paragraph{Face-space metric.}
CLIP was trained for image-text alignment rather than face recognition
and underestimates identity-specific erasure when the steered image
preserves pose and clothing but changes the face. We therefore use
ArcFace embeddings~\citep{deng2019arcface} $\phi$ and propose
\textbf{FaceAWUS}:
\[
  \mathrm{FaceAWUS}(s)
  =
  \mathbb{E}_{p\in\mathcal{T}}
  \left[
  1-\cos\!\left(\phi(g_p^s),\phi(g_p^0)\right)
  \right]
  -
  \mathbb{E}_{p\in\mathcal{A}}
  \left[
  1-\cos\!\left(\phi(g_p^s),\phi(g_p^0)\right)
  \right].
\]
where the first term measures how much the target face changes, and the
second measures how much adjacent faces change relative to the unmodified
generation. Since ArcFace is meaningful for identity-bearing face prompts, FaceAWUS is computed over adjacent identity prompts. Higher FaceAWUS is
better.
\paragraph{Perceptual metric.}
Finally, we measure perceptual selectivity using LPIPS:
\[
  \mathrm{LPIPS\mbox{-}gap}(s)
  =
  \mathbb{E}_{p\in\mathcal{T}}
  \left[
  \mathrm{LPIPS}(g_p^s,g_p^0)
  \right]
  -
  \mathbb{E}_{p\in\mathcal{N}}
  \left[
  \mathrm{LPIPS}(g_p^s,g_p^0)
  \right].
\]
This metric measures how much
more the target image changes than the adjacent and generic images. Higher LPIPS-gap indicates a more selective intervention.

Together, the three net utility metrics: AWUS, FaceAWUS and LPIPS-gap
give a complete picture across different representation spaces (CLIP, ArcFace, perceptual): the target concept should change
substantially, while adjacent and generic prompts should remain stable. 
Since a method may perform well in one space but poorly in another, reporting all three metrics helps prevent over-optimizing for any single notion of preservation.

\section{Model-internal identity adjacency}
\label{app:model_internal_adj}
Using frozen layer-9 SD~1.4 text-encoder name-token representations for 100 identities, we obtain the selected pairwise cosine similarities in Table~\ref{tab:model_adj_pairs}. This is separate from the CLIP-image-space adjacency analysis below and from SAE-based target-direction computation. The manually specified peer sets (Appendix~\ref{app:targets}) are not redefined by this post-hoc analysis. For 13 peer identities represented in the 100-identity corpus, mean reciprocal rank is 0.487.
\begin{table}[h]
\centering\small
\caption{\textbf{Selected model-internal identity relationships.} Cosine similarity in frozen SD~1.4 layer-9 text-encoder space.}
\label{tab:model_adj_pairs}
\begin{tabular}{lc}
\toprule
Pair & Cosine similarity \\
\midrule
Ronaldo--Messi & 0.460 \\
Obama--Biden & 0.394 \\
Musk--Zuckerberg & 0.401 \\
Cooper--Gosling & 0.421 \\
Johansson--Portman & 0.506 \\
\bottomrule
\end{tabular}
\end{table}
For example, Obama's four highest-ranking neighbors are Michelle Obama (0.574), Joe Biden (0.394), Kerry Washington (0.338), and David Cameron (0.332). These model-internal rankings complement rather than replace the semantically selected peer lists used in evaluation.

\section{Image-space identity adjacency}
\label{app:identity_adjacency}

To make the choice of adjacent identities not subjective and intuitive, we compute
pairwise image-space similarity between the unmodified baseline generations
of each identity. Higher similarity indicates that two identities occupy
nearby visual regions, motivating their use as adjacent identities. It is represented here by Figure~\ref{fig:identity_similarity}.

\begin{figure}[!t]
\centering
\includegraphics[width=0.75\linewidth]{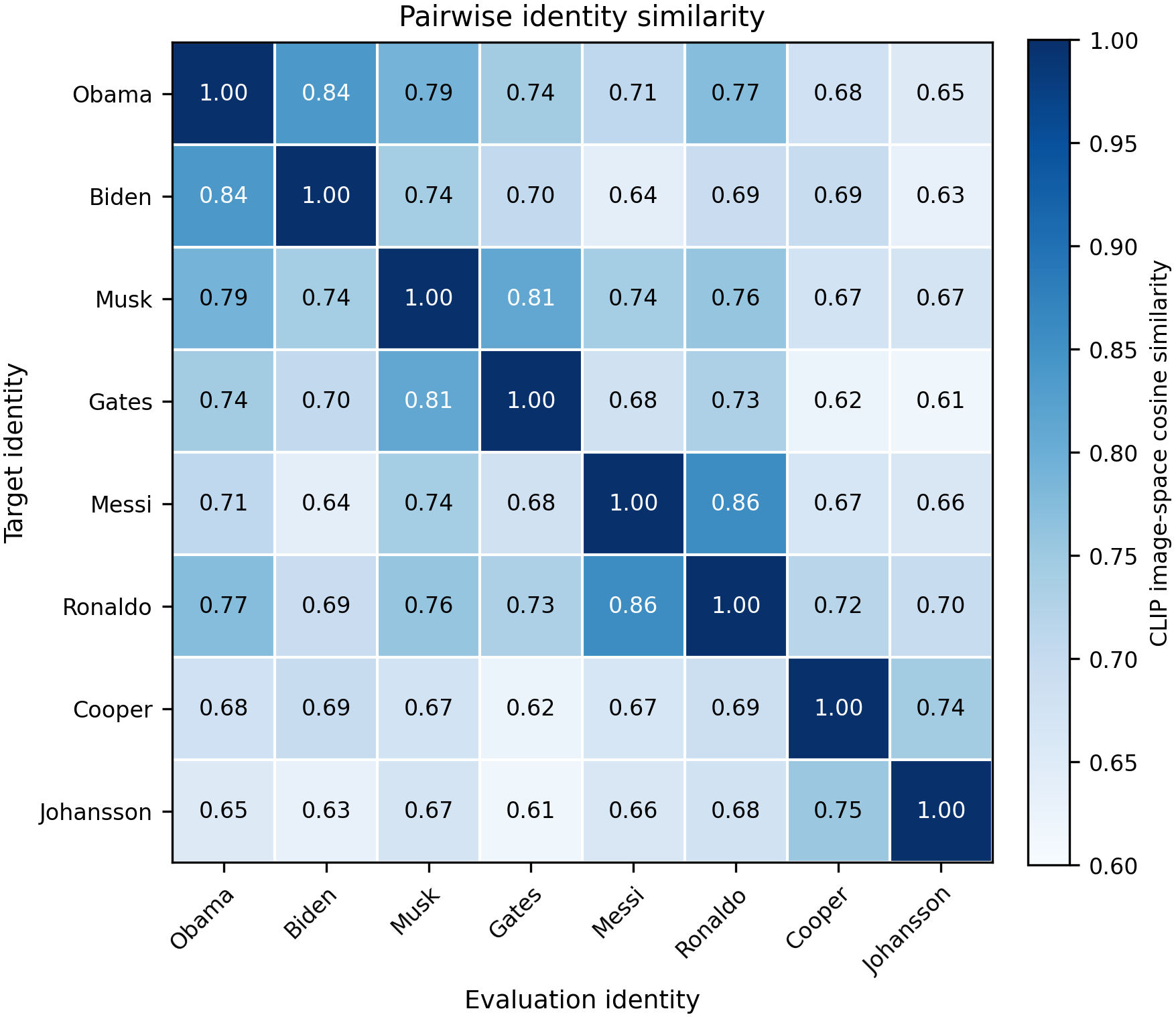}
\caption{%
  \textbf{Pairwise image-space identity similarity.}
  We compute cosine similarity between CLIP image embeddings of unmodified
  baseline generations for each identity. Higher values indicate visually
  closer identity prompts and provide an objective basis for selecting
  adjacent identities, e.g., Obama-Biden, Musk-Gates, Messi-Ronaldo, and
  Cooper-Johansson.
}
\label{fig:identity_similarity}
\end{figure}

\section{Additional trade-off analysis}
\label{app:face_tradeoff}

Figure~\ref{fig:face_scatter_app} visualizes the face-space trade-off
underlying FaceAWUS. A method that lies toward the top-left is ideal:
large target-identity change with small adjacent-identity drift.

\begin{figure}[!t]
\centering
\includegraphics[width=0.65\linewidth]{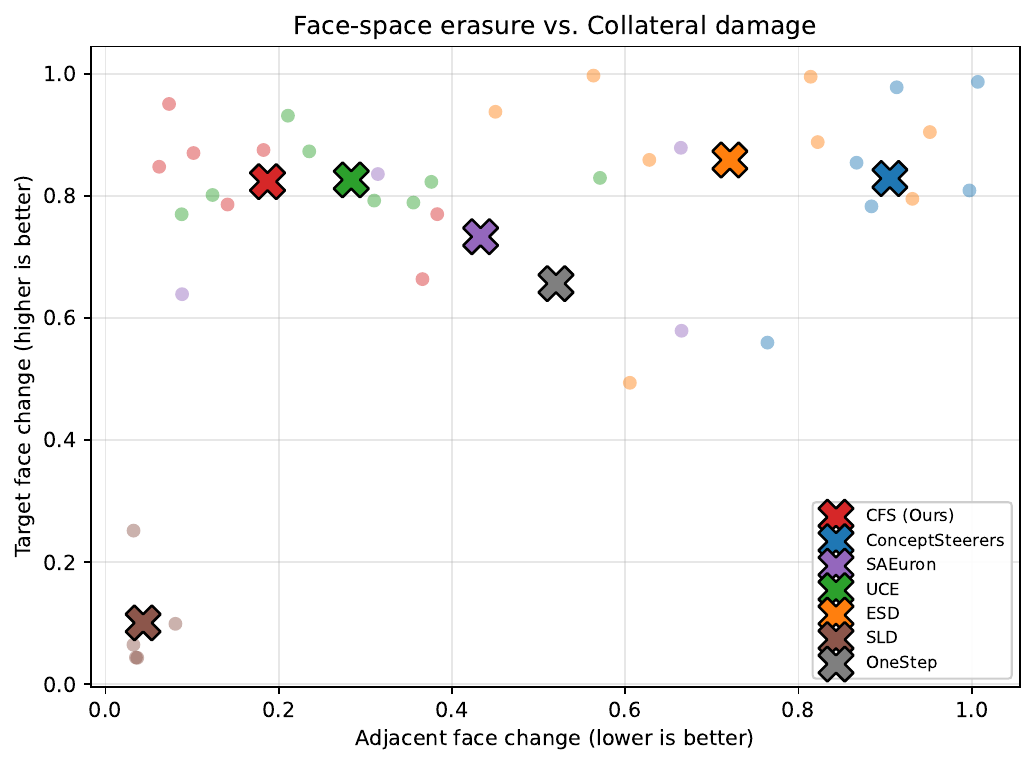}
\caption{%
\textbf{Face-space trade-off.}
CFS achieves high target-identity change with low adjacent-identity change,
while several baselines incur larger adjacent-face drift.
}
\label{fig:face_scatter_app}
\end{figure}

\section{Compositional and circumlocution examples}
\label{app:compositional}

Figure~\ref{fig:composability_adversarial_app} provides additional
qualitative examples showing that CFS can be composed across multiple target
identities and can also respond to indirect semantic references rather than
only explicit name tokens.

\begin{figure*}[!t]
\centering
\includegraphics[width=\linewidth]{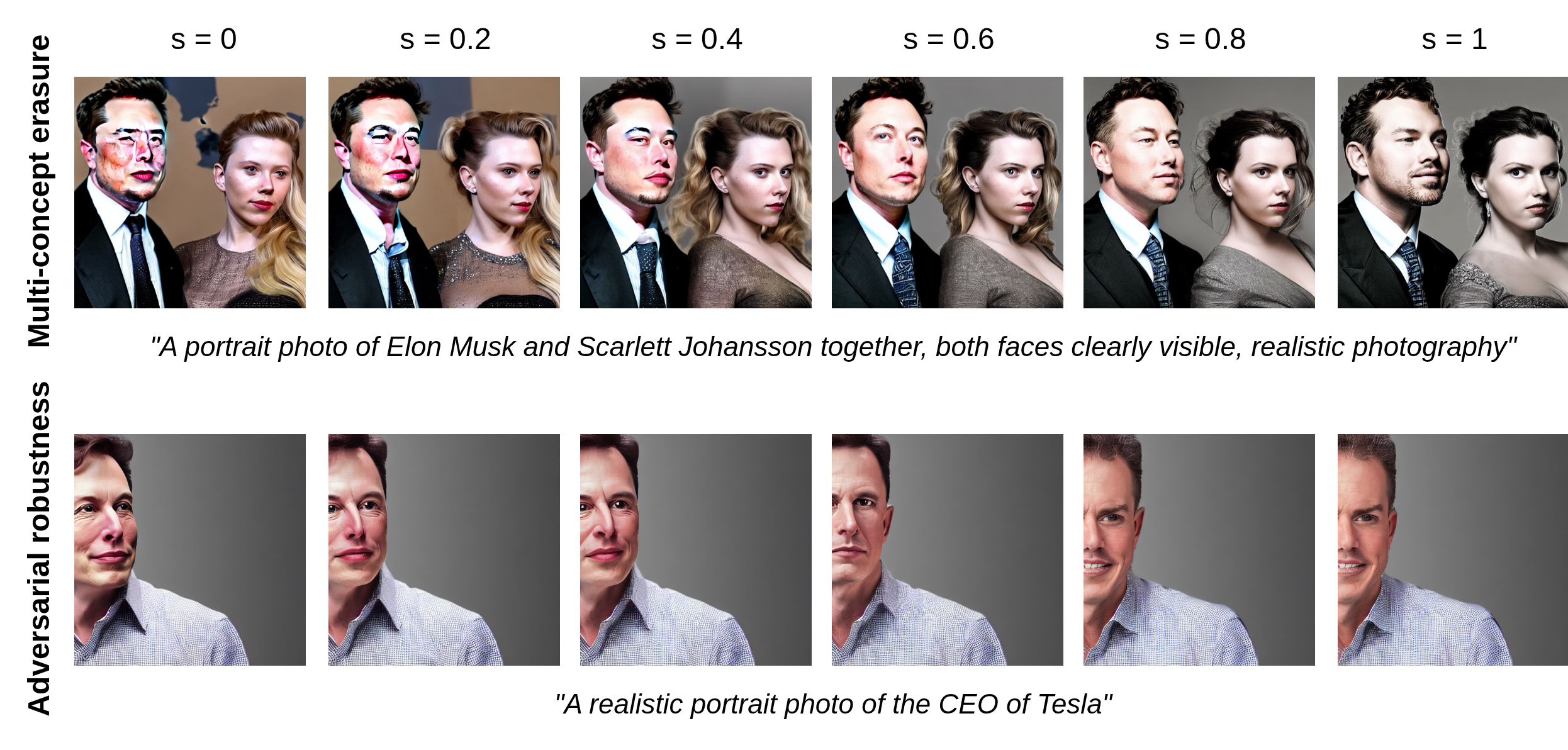}
\caption{%
\textbf{Compositional and semantic identity erasure.}
Top: CFS simultaneously erases Elon Musk and Scarlett Johansson in a single
generation. Bottom: CFS erases Elon Musk from a circumlocution prompt referring
to him as ``the CEO of Tesla''.
}
\label{fig:composability_adversarial_app}
\end{figure*}

\section{Token-level alignment}
\label{app:token_alignment}

Figure~\ref{fig:alpha_heatmap_app} shows where the CFS alignment signal
appears within a prompt. The signal concentrates on the identity-bearing name
tokens rather than generic style or composition tokens, supporting the claim
that CFS targets identity-specific features.

\begin{figure}[!t]
\centering
\includegraphics[width=0.8\linewidth]{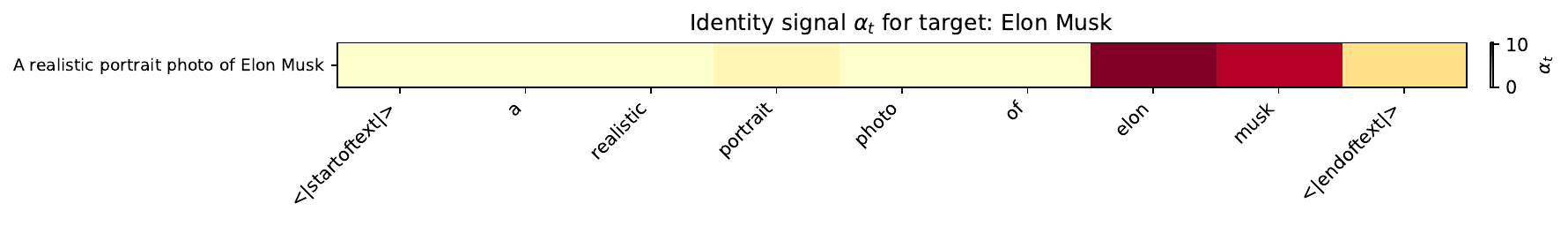}
\caption{%
\textbf{Per-token alignment signal for Elon Musk.}
The alignment signal concentrates on identity-bearing name tokens and remains
low on generic style or composition tokens.
}
\label{fig:alpha_heatmap_app}
\end{figure}

\section{Per-target Results}
\label{app:per_target}

\begin{figure}[!t]
\centering
\includegraphics[width=\linewidth]{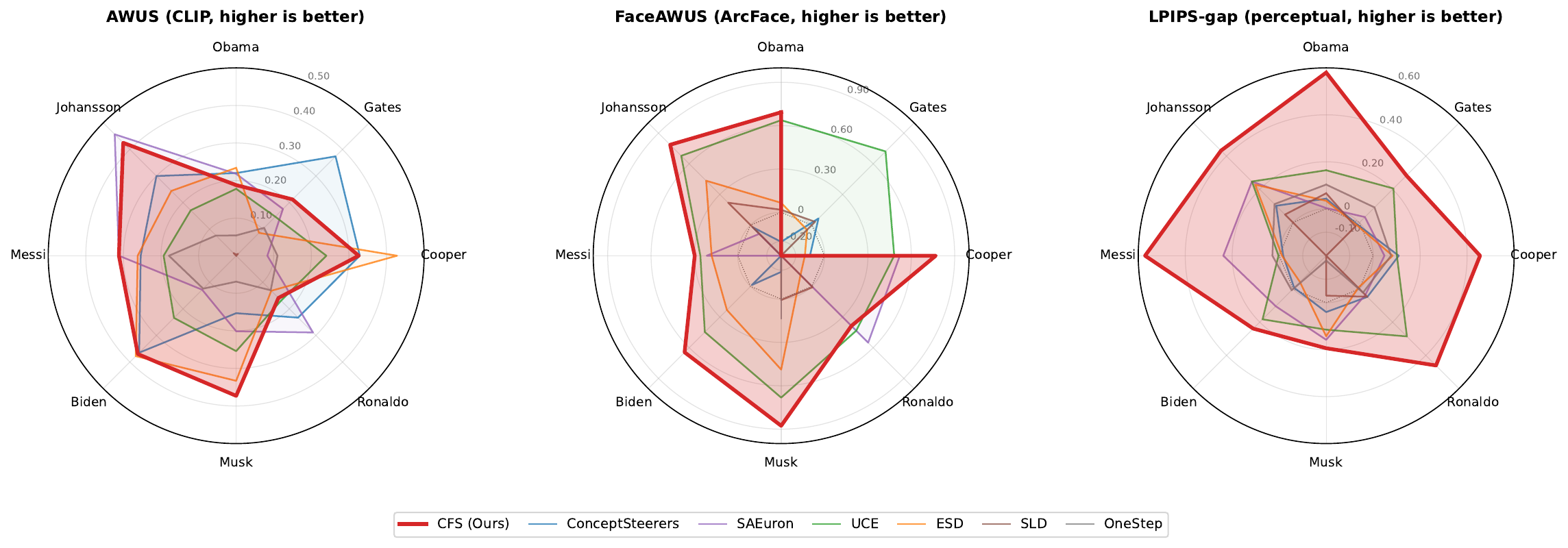}
\caption{\textbf{Per-target net utility across three metric spaces.} Each
panel is a radar over the 8 celebrity identities, with one polygon per
method. Larger area = better. CFS (red) achieves a balanced shape across
identities on AWUS, dominates the perimeter on FaceAWUS, and visibly
encloses every other method on every identity for LPIPS-gap. Missing
ArcFace face detections are floored at the inner radius (negative
extent); a polygon collapsing toward the centre on FaceAWUS indicates
a method whose erasure was so aggressive that ArcFace could not detect
a face in the steered image (e.g.\ SAEuron and OneStep on most targets).
The dashed circle at $0$ in the FaceAWUS and LPIPS-gap panels marks the
zero-utility threshold: methods with polygons inside this circle on a
given identity are doing more harm than good.}
\label{fig:radar_per_target}
\end{figure}

Figure~\ref{fig:radar_per_target} shows per-target net utility across
all three metric spaces. CFS achieves balanced performance across all
8 identities on AWUS, dominates on FaceAWUS, and encloses every other
method on LPIPS-gap. Methods whose polygons collapse toward the centre
on FaceAWUS (SAeUron, One-Step) performed such aggressive erasure that
ArcFace could not detect a face in the steered output.
\section{Runtime Comparison}
\label{app:runtime}

Table~\ref{tab:runtime} reports generation-time overhead. Since CFS modifies
the text-encoder representation once per prompt, rather than at every diffusion
step, so its overhead is independent of the number of denoising steps. CFS
achieves the best runtime among the evaluated unlearning methods, while
remaining essentially as fast as the unmodified SD\,1.4 baseline.

\begin{table}[h]
\centering
\caption{%
  \textbf{Runtime comparison.}
  Seconds per image on one A30 GPU using 50 DDIM steps, averaged over
  20 runs after 3 warm-up runs. CFS modifies the text-encoder representation
  once per generation, so its cost is independent of the number of diffusion
  steps.
}
\label{tab:runtime}
\begin{tabular}{lc}
\toprule
\textbf{Method} & \textbf{s/image} \,$\downarrow$ \\
\midrule
SD\,1.4 (no erasure)   & $9.52$ \\
\textbf{CFS (ours)}    & \textbf{9.50} \\
Concept Steerers       & 9.96\\
SAeUron                & 15.75 \\
One-Step               & 10.42 \\
UCE                    & 18.75 \\
SLD                    & 25.44 \\
ESD (incl.\ fine-tune) & 21.23 \\
\bottomrule
\end{tabular}
\end{table}


\section{Fixed-full-strength CLIP-space comparison}
\label{app:fixed_strength}
The main comparison allows each strength-controlled baseline its reported AWUS operating point while CFS uses $s=1$. For transparency, Table~\ref{tab:fixed_strength} reproduces the original separate supplementary table of CLIP-space scores at each method's full-strength setting. Strength parameters are method-specific and not numerically calibrated across approaches. Face-space metrics are not available at this matched nominal operating-point comparison for every method.
\begin{table}[h]
\centering\small
\caption{\textbf{Full-strength comparison in CLIP space.} Scores averaged across eight celebrity targets; $\mathrm{AWUS}=\mathrm{TER}-\mathrm{CDR}$.}
\label{tab:fixed_strength}
\begin{tabular}{lccc}
\toprule
Method & TER $\uparrow$ & CDR $\downarrow$ & AWUS $\uparrow$ \\
\midrule
CFS (ours) & 0.306 & \textbf{0.011} & \textbf{0.296} \\
Concept Steerers & \textbf{0.479} & 0.447 & 0.032 \\
SAeUron & 0.450 & 0.255 & 0.195 \\
UCE & 0.235 & 0.037 & 0.198 \\
ESD & 0.394 & 0.132 & 0.262 \\
SLD & 0.010 & 0.025 & $-$0.015 \\
One-Step & 0.438 & 0.337 & 0.101 \\
\bottomrule
\end{tabular}
\end{table}


\section{Reproducibility and Implementation Details}
\label{app:reproducibility}

\paragraph{Model, prompts, and concept directions.}
Our primary experiments use Stable Diffusion~1.4 with an intervention at layer~9
of its frozen CLIP text encoder. We construct 4,000 prompts derived from
Celeb-100, spanning 100 identities, and evaluate eight target identities;
for each target, the forget set contains prompts naming that target and the
retain set contains the remaining prompts. We evaluate four designated
adjacent identities per target and nine shared generic prompts. A domain-level
$k$-sparse autoencoder (SAE) is trained once on the corresponding text-encoder
activations and reused for all eight identity targets. Per-target directions
and retain centroids are computed from cached, no-gradient forget/retain
statistics; the diffusion-model parameters remain unchanged. Nudity and
artistic-style experiments use separately trained domain-level SAEs, rather
than assuming that the identity SAE transfers unchanged. The feature-selection
and projection equations appear in Section~\ref{sec:method}.

\paragraph{Controlled evaluations.}
For the main identity comparison, methods are evaluated with identical target
sets, prompts, random seeds, inference steps, and guidance settings. CFS uses
its full intervention strength $s=1$; other strength-controlled baselines use
their best evaluated AWUS operating points, and weight-editing baselines use
their target-specific edited checkpoints (Section~\ref{sec:setup}). The
additional face-specific evaluation uses ArcFace on detected faces, with
identity references obtained from unsteered Stable Diffusion~1.4 generations
rather than real celebrity photographs. The general-preservation evaluation
uses 10,000 COCO-30K captions with matching seeds and generation settings;
FID for the baseline and CFS is computed against the same corresponding COCO
reference images. The safety transfer study reports nudity detection rates on
I2P, Ring-a-Bell (K77, K38, K16), and MMA-Diffusion. The adjacency rankings
use frozen layer-9 name-token representations (found this layer to be the most useful based on ablation), rather than the trained SAE or
the generated-image evaluation metrics.

\paragraph{Computational accounting.}
The identity SAE requires approximately two hours of one-time training in the
reported setup; each target's cached statistics occupy approximately 925~KB.
The controlled runtime comparison uses one NVIDIA A30 GPU, 50 DDIM steps,
20 measured runs following three warm-up runs, and reports 9.50~s/image for
CFS versus 9.52~s/image for unmodified Stable Diffusion~1.4
(Appendix~\ref{app:runtime}).

\fi
\end{document}